\documentclass[10pt,twocolumn,letterpaper]{article}

\usepackage{cvpr}
\usepackage{times}
\usepackage{epsfig}
\usepackage[font=small,skip=5pt]{caption}
\usepackage{graphicx}
\usepackage{amsmath}
\usepackage{amssymb}
\usepackage{comment}
\usepackage{siunitx}
\DeclareMathOperator*{\E}{\mathbb{E}}
\DeclareMathOperator*{\argmin}{argmin} 
\newcommand{\mT}{\mathcal{T}}
\newcommand{\mL}{\mathcal{L}}
\newcommand{\hI}{\hat{I}}
\newcommand{\hp}{\hat{p}}

\newcommand{\bF}{\textbf{F}}
\newcommand{\bG}{\textbf{G}}

\graphicspath{{Figures/}}

\usepackage[pagebackref=true,breaklinks=true,letterpaper=true,colorlinks,bookmarks=false]{hyperref}

 \cvprfinalcopy 

\def\cvprPaperID{1495} 

\ifcvprfinal\fi
\begin{document}

\title{Guiding human gaze with convolutional neural networks}
\author{Leon A. Gatys\qquad
Matthias K\"ummerer\qquad
Thomas S. A. Wallis\qquad
Matthias Bethge \\
University of T\"ubingen\\
{\tt\small \{leon.gatys, matthias.kuemmerer, tom.wallis, matthias.bethge\}@bethgelab.org}}


\twocolumn[{%
\renewcommand\twocolumn[1][]{#1}%
\maketitle
\begin{center}
\includegraphics[width=1\textwidth]{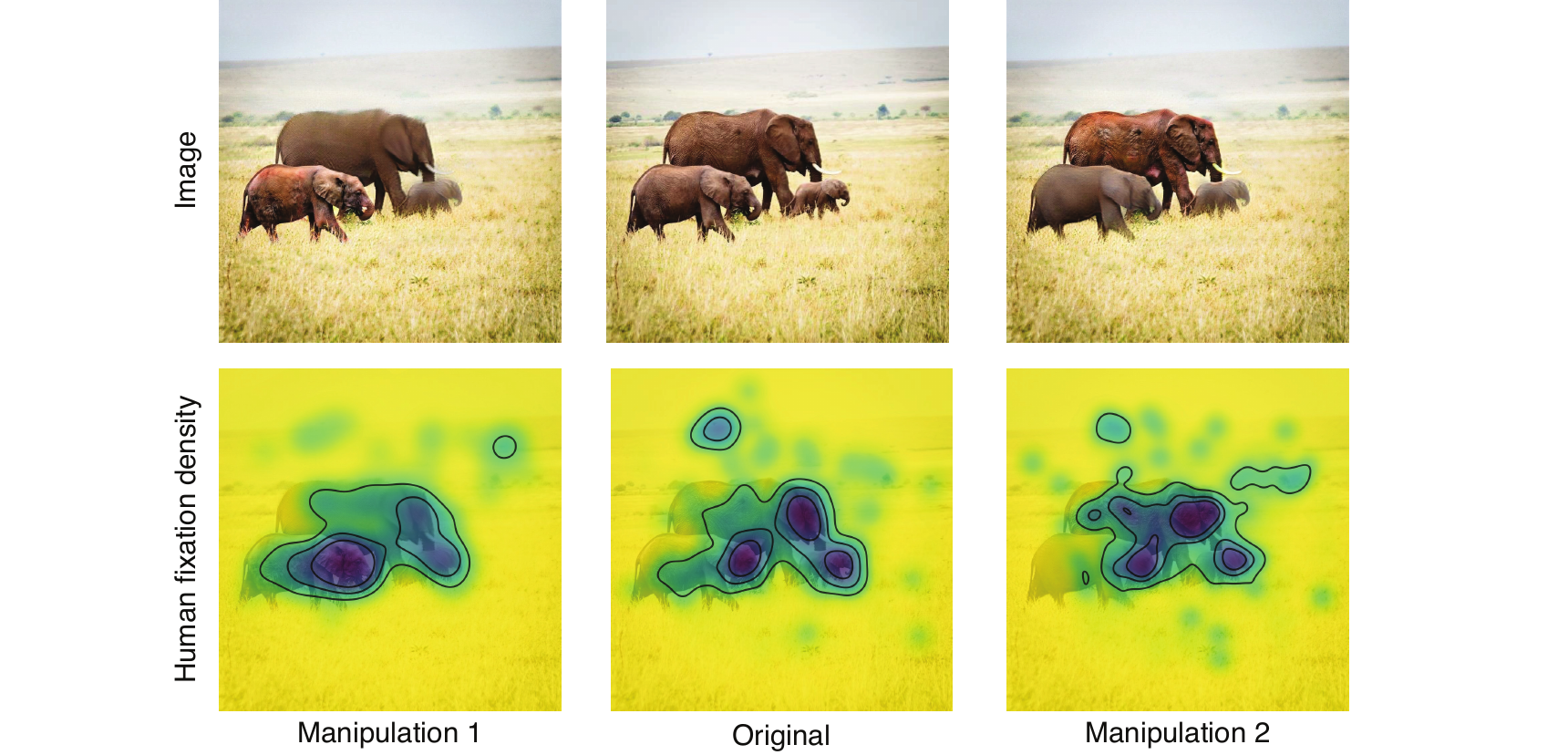}
\captionof{figure}{
Manipulating images to change where people look.
Images in the first and third column were generated by our model by transforming the original image shown in the second column.
The presented images are shown in the first row and the measured human fixation densities are shown in the second row.
The first manipulation aimed to make people fixate on the elephant in the foreground. 
The second manipulation aimed to make people fixate on the large elephant in the background.
Compared to the original image, the total probability to fixate on the target elephant increased by 0.17 (86\%), in the first image and 0.1 (28\%) in the second image.
}
\label{fig:Teaser}
\end{center}
}]


\begin{abstract}
The eye fixation patterns of human observers are a fundamental indicator of the aspects of an image to which humans attend. 
Thus, manipulating fixation patterns to guide human attention is an exciting challenge in digital image processing. 
Here, we present a new model for manipulating images to change the distribution of human fixations in a controlled fashion.
We use the state-of-the-art model for fixation prediction to train a convolutional neural network to transform images so that they satisfy a given fixation distribution.
For network training, we carefully design a loss function to achieve a perceptual effect while preserving naturalness of the transformed images.
Finally, we evaluate the success of our model by measuring human fixations for a set of manipulated images. 
On our test images we can in-/decrease the probability to fixate on selected objects on average by 43/22\% but show that the effectiveness of the model depends on the semantic content of the manipulated images.\footnote{Supplement at: bethgelab.org/media/uploads/gazeguide/Supplement.zip}
\end{abstract}

\section{Introduction}
Humans typically guide their visual attention selectively to different parts of an image and the spatial distribution of visual attention over an image strongly shapes perception.
Since human vision is foveated, the most important measurable correlate of visual attention is the position at which the fovea is fixated. 

Recently, models based on features from convolutional neural networks (CNNs) trained on object recognition have lead to major advances in predicting human fixation locations, also called saliency prediction \cite{kummerer_deep_2015, kummerer_understanding_2017, kruthiventi_deepfix:_2017,huang_salicon:_2015}. 
Furthermore, the same feature spaces allow to generate and manipulate images with respect to important perceptual properties such as objects \cite{nguyen_synthesizing_2016}, text \cite{nguyen_plug_2016}, image texture \cite{gatys_texture_2015} or artistic style \cite{gatys_image_2016}.
These image manipulations on perceptual variables are achieved by optimising perceptual loss functions defined in neural representations of CNNs trained on object recognition.

In this work, we aim to combine the recent success in predicting human fixation patterns with the advances in CNN-based image manipulation.
We explore to what extent we can use the information captured by a CNN trained on fixation prediction to inform image manipulations that change the distribution of human fixations (also called \emph{saliency map} \cite{kuemmerer_information-theoretic_2015, kummerer_saliency_2017}) of images in a controlled way.

This problem is challenging for two main reasons.
First, the saliency map of an image contains far less information than the image itself. 
Thus, there exists a myriad of images that would satisfy a given target saliency map, many of which are quite unnatural.
Second, we find that the state-of-the-art model for human fixation prediction \cite{kummerer_understanding_2017} is prone to adversarial examples, similar to other CNN-based prediction models \cite{szegedy_intriguing_2013, cisse_houdini:_2017}. 
That means, one can apply changes to an image that are imperceptible for humans but make the image satisfy an arbitrary target saliency map for the prediction model.
We address these challenges by carefully designing a loss function to preserve identity and naturalness of the transformed images.
Moreover, we reduce the flexibility of the admissible image transformation by training a fixed CNN architecture to manipulate the saliency map of a large dataset of images such that it cannot overfit on adversarial noise for specific examples

Finally, we conduct a behavioural experiment to evaluate the success of our method.
We construct a set of manipulated images and measure human fixation patterns in response to these images. 
Analysing the results, we find that the effectiveness of our method depends on the semantic content of the image but in most cases we successfully change the fixation patterns in the desired fashion. 


\section{Related work}
Several previous studies aim to manipulate images with respect to saliency (for review see \cite{mateescu_visual_2015}).
However, the nature of the image manipulations as well as the measure of image saliency varies.
Image saliency is typically measured by some hand-crafted model (e.g. \cite{itti_model_1998}) and the image manipulations aim to directly change the features that are used for saliency prediction (e.g. colour \cite{mateescu_attention_2014, nguyen_image_2013}, frequency bands \cite{su_-emphasis_2005} and luminance-contrast \cite{einhauser_does_2003, vig_learned_2011}  or a mixture of them \cite{wong_saliency_2011, hagiwara_saliency-based_2011}). 
There are often some additional constraints to preserve naturalness of the output image (e.g. total limits on the manipulation \cite{wong_saliency_2011, su_-emphasis_2005} or only to use colours from within the image \cite{mechrez_saliency_2016} or of similar objects from a database \cite{nguyen_image_2013}).
In contrast to our work, none of the previous studies aimed to learn a general saliency-manipulating image transformation directly from a data-driven state-of-the art model for saliency prediction.

Similar to us, a number of studies have used pre-trained CNN features to synthesise or manipulate images with regard to perceptual properties (for review see \cite{gatys_texture_2017}).
Successful examples include attribute-based image synthesis \cite{simonyan_deep_2013, nguyen_synthesizing_2016, nguyen_plug_2016} and manipulation \cite{upchurch_deep_2016}, texture synthesis and style transfer \cite{gatys_texture_2015, gatys_image_2016}.
Technically most closely related to our work are studies that train a CNN to transform one image into another image.
Example tasks include image generation from segmentations \cite{isola_image--image_2016, shrivastava_learning_2016, chen_photographic_2017}, style transfer \cite{johnson_perceptual_2016, ulyanov_texture_2016, li_precomputed_2016, zhu_unpaired_2017} or superresolution \cite{johnson_perceptual_2016, ledig_photo-realistic_2016, sajjadi_enhancenet:_2016}.
Note that in contrast to many other image to image translation tasks (e.g. \cite{isola_image--image_2016}), there exists no ground truth data in pixel space for our saliency manipulation task.


%

\section{Saliency manipulation method}
\begin{figure}
\includegraphics[width=1\linewidth]{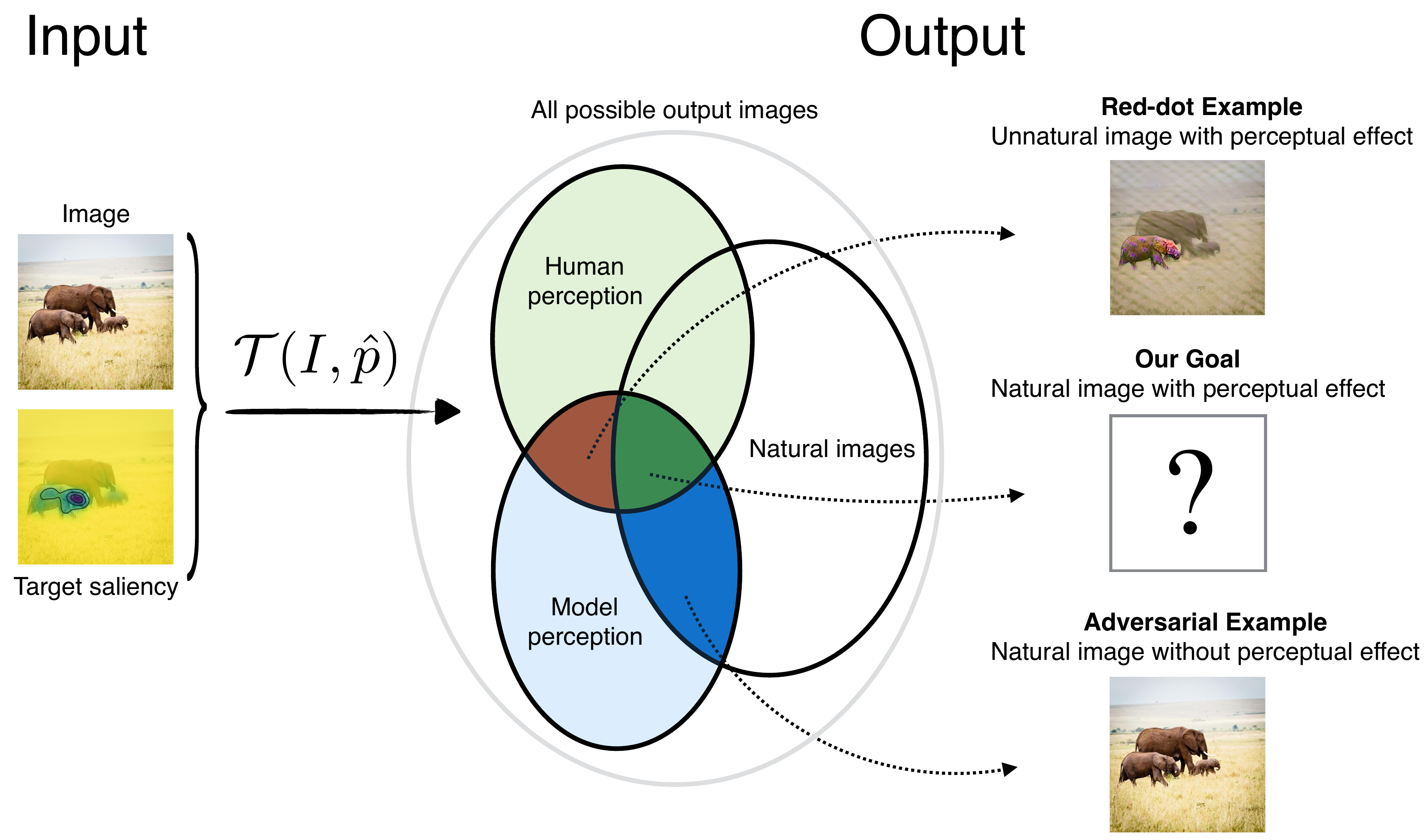}
\caption{Conceptual setup:
An input image and a target saliency map are transformed into an output image.
Output images can have the desired saliency effect for humans (light green area, `Human perception') and for the saliency model (light blue area, `Model perception') and can be natural or artificial.
The transformation is guided by model perception but our goal is to produce natural images that affect human perception in the desired way (dark green area).
Artificial images can easily achieve desired saliency effects for humans (red area).
Adversarial examples can easily achieve desired saliency effects for the model while maintaining naturalness (dark blue area). 
}
\label{fig:TransSetup}
\end{figure}

The basic problem we are addressing can be formulated as follows:
We want to find a transformation $\mathcal{T}$ that maps an image $I(x, y)$ and a target fixation distribution $p_{t}(x, y)$ to a new image $I_{t}(x, y)$ such that, when observing $I_{t}(x, y)$, human fixation patterns satisfy the specified target distribution $p_{t}(x, y)$ (domain labelled `Human perception' in Fig. \ref{fig:TransSetup}).

However, since collecting human fixation patterns is expensive, we cannot directly guide our search for an appropriate image transformation by human behavioural data.
Instead, we need to use a model that predicts human fixations and can be easily evaluated on new images.

\subsection{Predicting human fixation patterns}
To model the density of human fixations, we use the most recent DeepGaze model \cite{kummerer_understanding_2017}, a saliency prediction model that achieves state-of-the-art performance in fixation prediction as evaluated on the MIT300 saliency benchmark \cite{mit-saliency-benchmark}.
The model takes an image as input  and outputs a probability distribution over pixels indicating the fixation probability at each image location:
\begin{align}
DG(I(x,y)) = p(x,y)
\end{align}

To predict fixation densities, the model uses the feature spaces of the VGG-19 Network \cite{simonyan_very_2014} that was trained on the ImageNet object recognition challenge \cite{russakovsky_imagenet_2014}.
In particular, it uses features from layers conv5\_1,  relu5\_1, relu5\_2, conv5\_3, relu5\_4 giving a three-dimensional tensor with 2560 ($5 \times 512$) channels.
It computes a point-wise non-linear combination of the VGG features using a four-layer readout network with $1 \times 1$ rectified convolutions in each layer to produce a single output channel.
This channel is up-sampled by a factor $8$ and blurred with a gaussian kernel to regularise the predictions.
Finally, a center-bias is added to the channel to model the prior distribution over fixations.
The result, $S(x, y)$, is converted into a probability distribution over the image using a soft-max function over spatial positions:
\begin{align}
\label{eq:sm}
  p(x, y) = \frac{\exp(S(x, y))}{\sum_{x, y} \exp(S(x, y))}
\end{align}
Here, we omit the center-bias of DeepGaze when using it for saliency manipulation, since we want to inform our image transformations only with image dependent saliency information.

%


\subsection{Saliency loss}
To manipulate the saliency map of an image $I$ to match a target saliency map $p_t$, we transform it by $\mT$ to generate a new image $\hI$: $\hI = \mT(I,p)$.
Next, we compute the saliency map $\hp$ of the transformed image: $\hp=DG(\hI)$.
To measure the success of the image transformation, we compute the KL-Divergence from the saliency map of the transformed image to the target saliency map:
\begin{align}
\mathcal{L}_{sal}= \sum_{x,y}p_t(x,y)\log\left(\frac{p_t(x,y)}{\hp(x,y)}\right)
\end{align}

Importantly, we are limited how well we can manipulate where people look by the agreement of our saliency model with human fixations. In fact, we can only directly search for images that affect the perception of DeepGaze (domain labelled `Model perception' in Fig. \ref{fig:TransSetup}) but not for images that affect human perception. Still, we want to find images that not only affect the model perception but also human perception (intersection between `Model perception' and `Human perception' domains in Fig. \ref{fig:TransSetup}).

\subsection{Preserve naturalness}
Manipulating the saliency map of images in any way is not necessarily useful. 
For example, guiding the observer's attention by placing a bright red dot in the image would hardly be considered an interesting image manipulation. 
Similarly, there exist many transformations of the input image that strongly distort the image to match the target saliency map. 
These images can produce a perceptual effect for both the saliency model and humans, but lie outside the domain of natural images (red area of intersection between `Model perception' and `Human perception' domains Fig. \ref{fig:TransSetup}) and thus the transformations that produce them are of limited use.
Here we employ several measures to ensure that we are searching for image transformations that stay in the domain of natural images and somewhat close to the input image in particular.

\subsubsection{Feature loss}
We aim to preserve the overall structure and content of the input image but leave flexibility for identity-preserving image transformations. 
To that end, we penalise the distance of the transformed image to the input image in a feature space provided by a deep layer of the VGG network.
Say $\bF_\ell(I)$ is the feature representation of an image $I$ in layer $\ell$ of the VGG-network. Each column of $\bF_\ell(I)$ is a vectorised feature map and thus $\bF_\ell \in \mathbb{R}^{M_\ell(I)\times N_\ell}$ where $N_\ell$ is the number of feature maps in layer $\ell$ and $M_\ell(I) = H_\ell(I)\times W_\ell(I)$ is the product of height and width of each feature map.
We measure the mean-squared error between the feature representation of the input image $I$ and the transformed image $\hI$:
\begin{align}
\mathcal{L}_{feat} = \frac{1}{N_{\ell}M_{\ell}(I)} \sum_{ij}\left(\bF_{\ell}(\hat{I})-\bF_{\ell}(I)\right)_{ij}^2
\end{align}
This is the same as the well-known content loss from Neural Style Transfer \cite{gatys_image_2016}, only that here we use layer relu5\_2 instead of layer relu4\_2.
This choice worked well in our experiments but we did not exhaustively compare between different choices of feature representations.

\subsubsection{Texture loss}
We also want to preserve the overall appearance of the input image and its low-level structure without enforcing its exact reconstruction. For that purpose, we employ a texture loss \cite{gatys_texture_2015} that measures the difference between the texture of the transformed image and the texture of the output image:
\begin{align}
\label{eq:tex}
\mathcal{L}_{tex} &= \sum_\ell w_\ell E_\ell \\
E_\ell &= \frac{1}{N_\ell^2}\sum_{ij}{\left(\bG_\ell(\hI) - \bG_\ell(I)\right)_{ij}^{2}}
\end{align}
where $\bG_\ell(I)=\frac{1}{M_\ell(I)}\bF_\ell(I)^T\bF_\ell(I)$ is the Gram Matrix of the feature maps in layer $\ell$ in response to image $I$.
We include Gram Matrices from layers relu1\_1, relu2\_1, relu3\_1, relu4\_1, relu5\_1 with equal weights to model the texture of the input image.
For both the texture and the feature loss, we used the same normalised VGG-network \cite{gatys_texture_2015} that is also used by DeepGaze \cite{kummerer_understanding_2017}.
\subsubsection{Adversarial loss}
Finally, we employ a patch-based conditional adversarial loss \cite{isola_image--image_2016, shrivastava_learning_2016, li_precomputed_2016}.
An adversarial loss \cite{goodfellow_generative_2014} is an adaptive loss term aiming to correct for systematic differences between the input and the transformed images.
In the adversarial loss, a discriminator learns to discriminate between image patches of transformed and input images and is jointly optimised with the image transformation. 
If the image transformation produces images whose patches are systematically different from the input images, the discriminator can learn this difference and inform the image transformation to correct for it.

For the discriminator we used the implementation of the patch-based adversarial loss by \cite{isola_image--image_2016} with receptive field size of 70 px. 
We used the LSGAN \cite{mao_least_2016} objective for improved training stability and thus, the following term is added to the loss function for the image transformation:
\begin{align}
\label{eq:advt}
\mL_{adv} = \sum_{x,y}(1 - D(\hI)(x,y))^2
\end{align}
where $D(I)(x,y)$ denotes the single channel output of the discriminator in response to image $I$.
At the same time, the discriminator CNN is optimised to distinguish between input and transformed images:
\begin{align}
\label{eq:advd}
\mL_{D} = \frac{1}{2}\sum_{x,y}\left(D(\hI)(x,y)^2 + (1 - D(I)(x,y))^2 \right)
\end{align}
This patch-based conditional adversarial loss can also be understood as a texture loss. 
However, in our previously described texture loss the loss function is fixed and based on the pre-trained VGG features, whereas the adversarial loss function is adaptive and trained from scratch.

\subsubsection{Avoid adversarial transformations}
The total loss function we aim to minimise with respect to the image transformation is: 
\begin{align}
\label{eq:loss}
\mL_{total} = \lambda_{s} \mL_{sal} + \lambda_{f} \mL_{feat} + \lambda_{t} \mL_{tex} + \lambda_{a} \mL_{adv} 
\end{align}
For a given input image and target saliency map, the most flexible approach to minimise this loss function is to directly optimise the pixels of the input image, as was previously done for CNN-based image generation \cite{simonyan_deep_2013, mahendran_understanding_2014, gatys_texture_2015, yosinski_understanding_2015}.
Unfortunately though, we find that optimising the image directly leads to adversarial examples \cite{szegedy_intriguing_2013}: transformed images that are indistinguishable from the input images for humans, but match the required target saliency map predicted by DeepGaze.
Thus, they remain natural images and generate a perceptual effect for the model but do not generate a perceptual effect for humans (dark blue area of intersection between `Model perception' and `Natural images' domains Fig. \ref{fig:TransSetup})

Adversarial examples can exist for any model whose prediction is based on different image information than human perception.
Since our saliency prediction model is not perfect, the existence of adversarial examples is not surprising.
Because the saliency loss is the only term that encourages the transformed image to be different from the input image, an image transformation that generates adversarial examples for DeepGaze will minimise our total loss function (Eq. \ref{eq:loss}).
Thus, to avoid adversarial image transformations, we need to constrain the class of image transformations over which we optimise. 

\begin{figure}
\includegraphics[width=1\linewidth]{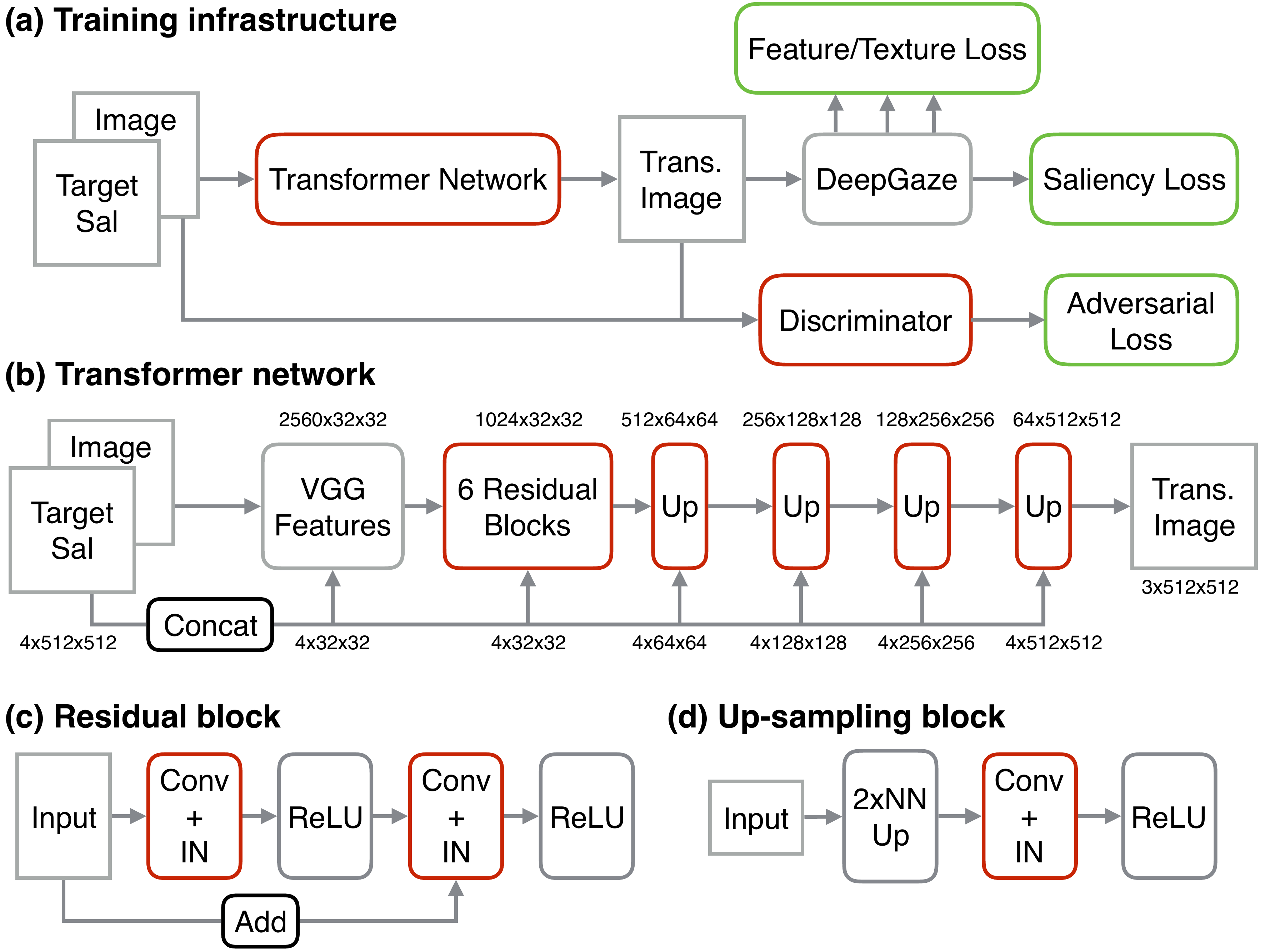}
\caption{
\textbf{(a)} Training infrastructure. Modules with trainable parameters are red, contributions to the loss function are green. Grey modules are not optimised. 
\textbf{(b)} Network architecture. VGG features are computed from the input image and in subsequent stages transformed and up-sampled to generate the transformed image. The input image and target saliency map is concatenated with the feature maps at several stages of the transformer network.
\textbf{(c)} Residual block: Two convolutional layers with InstanceNorm (IN) and ReLU non-linearities. The input is added before the second ReLU.
\textbf{(d)} Up-sampling block: Nearest-neighbour up-sampling followed by a convolutional layer with IN and ReLU.
}
\label{fig:ImgTrans}
\end{figure}

\subsection{Image Transformation}
We define the class of admissible image transformations by parameterising $\mT$ as a CNN with parameters $\theta$.
Furthermore, we require that the same transformation simultaneously minimises our loss function for a large set of images (Fig. \ref{fig:ImgTrans}(a)).
Hence, searching for the optimal image transformation $\mT_{opt}(I, p_t, \theta_{opt})$ means training a CNN to minimise the loss function (Eq. \ref{eq:loss}) for many input images $I$ and target saliency maps $p_t$:
\begin{align}
\label{eq:opt}
\theta_{opt} = \argmin_{\theta} \E_{I, p_t}\left[\mL_{total}(I, p_t, \theta)\right]
\end{align}
During training, input images and target saliencies are passed to the transformer network, which produces output images.
The output image is passed to DeepGaze. 
The VGG features of DeepGaze are used to compute the feature and texture losses.
The final output of DeepGaze is used to compute the saliency loss (Fig. \ref{fig:ImgTrans}(a)).
At the same time, the input and the transformed image are passed to the discriminator network.
The discriminator computes the adversarial loss for the transformer network (Eq. \ref{eq:advt}) and optimises its own discrimination target (Eq. \ref{eq:advd}) (Fig. \ref{fig:ImgTrans}(a)).

\subsubsection{Transformer network}

Since existing CNN architectures for image-to-image mapping \cite{johnson_perceptual_2016, isola_image--image_2016} did not perform well on our task (Fig. \ref{fig:NetTrain}), we developed a new CNN architecture (Fig. \ref{fig:ImgTrans}(b)).
First, the input image is transformed into the 2560 VGG feature channels used by DeepGaze to predict the saliency map.
This feature space extracts the rich image information with large receptive fields that DeepGaze uses for saliency prediction.
Its spatial dimensions are 16 times smaller than that of the input image.
At that resolution, the image is transformed using 6 blocks of residual layers \cite{johnson_perceptual_2016, he_deep_2016} that have 1024 feature channels.
Each residual block consists of two stages of convolutional layer with kernel size $3\times3$ (Conv), InstanceNorm (IN) \cite{ulyanov_improved_2017} and rectifying-linear unit (ReLU).
The input to the residual block is added before the last ReLU (Fig. \ref{fig:ImgTrans}(c)).
Next, we have 4 up-sampling stages that increase each spatial dimension by a factor of 2 and decrease the number of channels by a factor of 2 (Fig. \ref{fig:ImgTrans}(b)).
Each up-sampling stage consists of nearest neighbour up-sampling (NN Up), Conv, IN and ReLU (Fig. \ref{fig:ImgTrans}(d)).
At each stage of the processing hierarchy, we want the transformer network to have access to the input image and target saliency map to inform the transformation and allow the preservation of low-level information.
Therefore, we down-sample the 4 input channels (RGB image and target saliency map) to the respective size and concatenate them with the feature output of the network at each processing stage (Fig. \ref{fig:ImgTrans}(b)).
Thus, the first residual block receives 2564 (2560+4), the first up-sampling block 1028 (1024+4) and the second up-sampling block 516 (512+4) input channels and so on.
In that way we have a powerful transformer architecture built on the rich object-based VGG features that can also preserve all low-level image information.
In the following we call this architecture `feature-guided transform' (FGTransform).

\subsubsection{Network training}
\label{sec:NetTrain}
We use the MSCOCO training images \cite{lin_microsoft_2014} to train our network on saliency manipulation.
The images are spatially resized to 512x512 pixels, and pre-processed for the VGG-network (transformed to BGR, subtraction of channel mean, and scaled by 255) \cite{simonyan_very_2014}.

We have to generate target saliency maps for each training sample.
We hypothesise that there is a class of `natural saliency maps' that arise from the set of natural images.
We tried to construct natural target saliency maps by changing the original saliency map of the input image in either of two ways.

\begin{figure}
\includegraphics[width=1\linewidth]{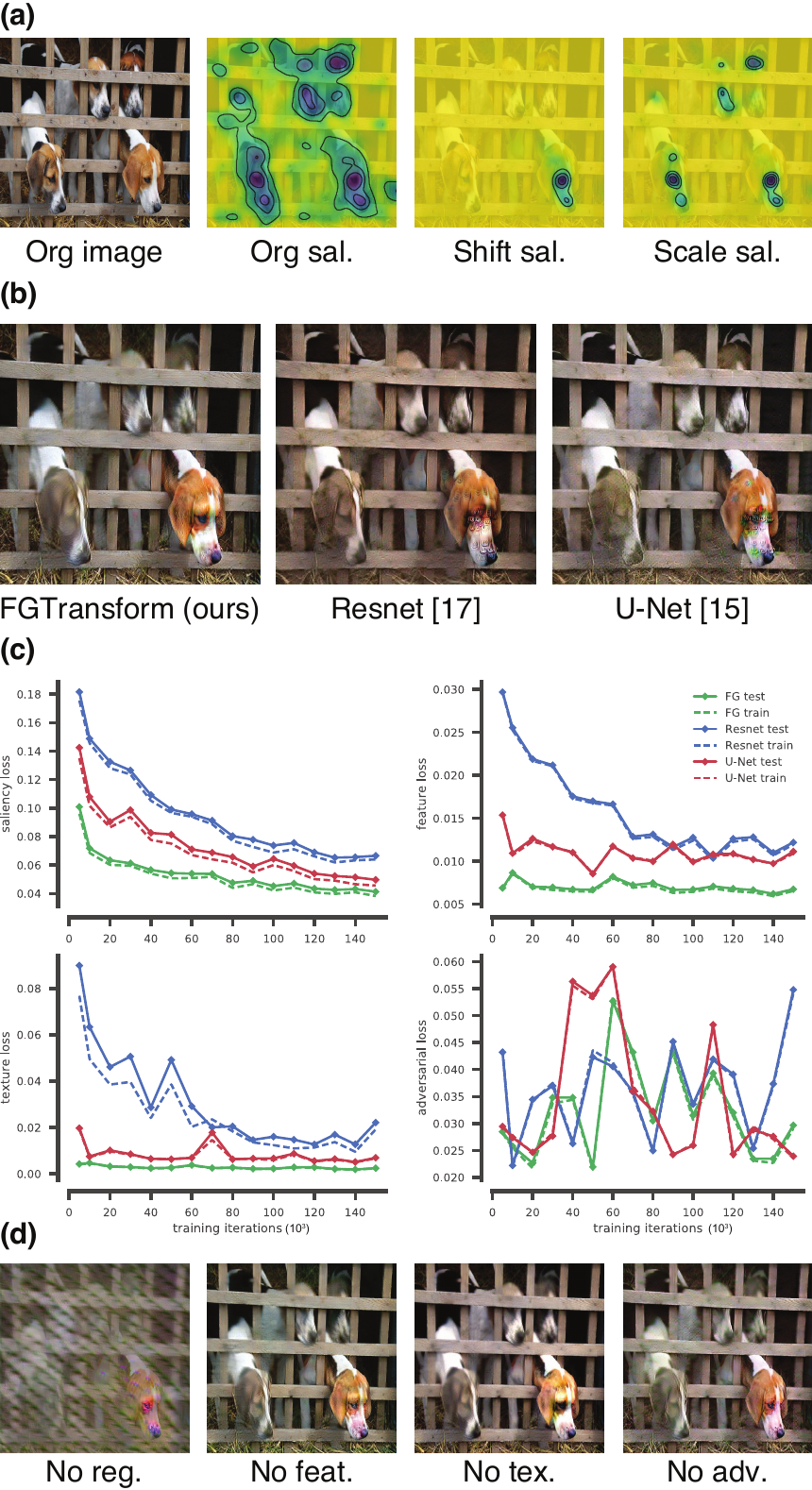}
\caption{
\textbf{(a)} Target saliency maps are either created by in-/decreasing the saliency of particular objects (third column) or globally scaling the saliency map (fourth column).
\textbf{(b)} Existing image-to-image networks generate artificial distortions in our task.
\textbf{(c)} Our feature-guided architecture minimises all parts of the loss function better than existing architectures. Results shown for training on local saliency shifts. Results for global saliency scaling look similar and can be found in the Appendix.
\textbf{(d)} Ablation studies. Without regularisation output images are very distorted (first column). Leaving out other parts of the loss function has subtle but noticeable effects.
Images best viewed with zoom.
}
\label{fig:NetTrain}
\end{figure}

In the first manipulation, we add a constant to a local region of the un-normalised saliency map:
\begin{align}
  S_t(x, y) = S(x, y) + k_{sh} M(x,y) 
\end{align}
Here, $k_{sh}$ is the constant shift to the saliency map and $M(x,y)$ denotes a mask that defines the local region which will be changed. 
Intuitively, this manipulation corresponds to increasing (for $k_{sh}>0$) or decreasing (for $k_{sh}<0$) the saliency of local parts of the image.
We sample the masks $M(x,y)$ from the object segmentation labels of the COCO dataset. Each mask is blurred with a gaussian kernel to keep the target saliency maps smooth.
Thus, during training the network learns to in-/decrease the saliency of annotated objects and people in the training set (Fig. \ref{fig:NetTrain}(a), third column).

For the second manipulation, we globally scale the saliency map with a constant factor:
\begin{align}
  S_t(x, y) = k_{sc} \times S(x, y) 
\end{align}
Here $k_{sc}$ denotes the constant scaling factor that is applied to the saliency map.
Intuitively, this manipulation changes the consistency of the fixation patterns.
It either increases the clustering of the fixations on few, very salient regions (for $k_{sc}>1$) or spreads the fixation patterns more uniformly over the image (for $0<k_{sc}<1$).
The factor $k_{sc}$ can also be thought of a temperature parameter that in-/decreases the entropy of the fixation distribution (Fig. \ref{fig:NetTrain}(a), fourth column).

In both cases, the target saliency distribution $p_t$ is obtained by normalising $S_t$ using the soft-max function (Eq. \ref{eq:sm}).
During training, we generate new target saliency maps on-the-fly for each training sample by either locally shifting all annotated objects in the image by a constant $k_{sh} \in [-4,4]$ or globally scaling by a factor $k_{sc} \in [0.5, 2]$. 
We always used the log saliency map during training, so the input to the transformer network is $\{I, \log(p_t)\}$.

Further details of the training procedure can be found in the Appendix.

\section{Results}
\subsection{Visual inspection}
We compare the training results between our FGTransform and the previously published `Resnet-9' and `U-Net' architectures \cite{johnson_perceptual_2016, isola_image--image_2016} (Fig. \ref{fig:NetTrain}(b),(c)).
For the `Resnet-9' and `U-Net' architectures, we observed problems with unnatural distortions in the transformed images (Fig. \ref{fig:NetTrain}(b), head of the salient dog in second and third column).
Interestingly, these distortions appear to resemble scribbled text, which is one of the main features driving human fixations \cite{kummerer_understanding_2017}.
Thus, it makes sense to put text at locations that should have high saliency, but this does not preserve the naturalness of the image well.
Ideally though, the transformation can increase the saliency of the features in the input image rather than always putting text scribbles in the corresponding location.
That is why we designed our own feature-guided architecture hoping that a better-suited network can learn a more image-dependent transformation.
We did not encounter similar problems with artificial distortions with our architecture (Fig. \ref{fig:NetTrain}(b), first column) indicating that it can learn a more image-dependent transformation.
Model comparisons for all stimuli used in the behavioural study in section \ref{sec:behave} can be found in the Supplement.

\subsection{Quantitative evaluation}
To compare the models quantitatively, we computed the training and test error over the training iterations for each saliency manipulation.
We sampled 1000 training images from the COCO training set and 1000 test images from the COCO validation set that were not used during training of the model.
As during training, we randomly sampled either $k_{sh}$ from $[-4,4]$ or $k_{sc}$ from $[0.5,2]$ to generate the target saliency for each image. 
We find that the training and test error are on a similar level and even after 150k training iterations there is no sign of overfitting (Fig. \ref{fig:NetTrain}(c)).
Also, when inspecting the transformed images we were unable to tell the difference between training and test images.
Furthermore, our network architecture consistently leads to smaller loss values than the compared `Resnet-9' and `U-Net' architectures (Fig. \ref{fig:NetTrain}(c)).
This is true for the saliency loss as well as the regularisation losses meaning that our model provides a better solution to the problem and not only a better trade-off between naturalness and saliency manipulation.

\subsection{Ablation studies}
Finally, we re-trained our FGTransform while setting different parts of the loss function equal to zero.
When only training on the saliency loss ($\lambda_{f}, \lambda_{t}, \lambda_{a}=0$), the transformed images are strongly distorted (Fig. \ref{fig:NetTrain}(d), first column).
When leaving out only one of the regularisation losses (either of $\lambda_{f}, \lambda_{t}, \lambda_{a}$ equal to 0), the differences are  more subtle.
We see that in each case, the images are slightly more distorted than for the full loss function \ref{fig:NetTrain}(d)).
Looking at many examples, we found that the addition of every part of the loss function increased the perceptual quality of the transformed images. 
Nevertheless, the perceptually optimal trade-off between the different regularisation terms is hard to determine since our loss function is only a rough quantitive measure of perceptual quality.
Ablation results for all stimuli used in the behavioural study in section \ref{sec:behave} can be found in the Supplement.

\section{Behavioural study}
\label{sec:behave}
We have developed an image transformation that can manipulate arbitrary input images to change the saliency prediction of DeepGaze while preserving naturalness. 
However, the existence of adversarial examples illustrates that model prediction and human perception can be quite different for images optimised with respect to the model.
Thus, to evaluate our work it is vital to measure human fixation patterns in response to images generated with our model.
\subsection{Stimulus generation}
\begin{figure}
\includegraphics[width=1\linewidth]{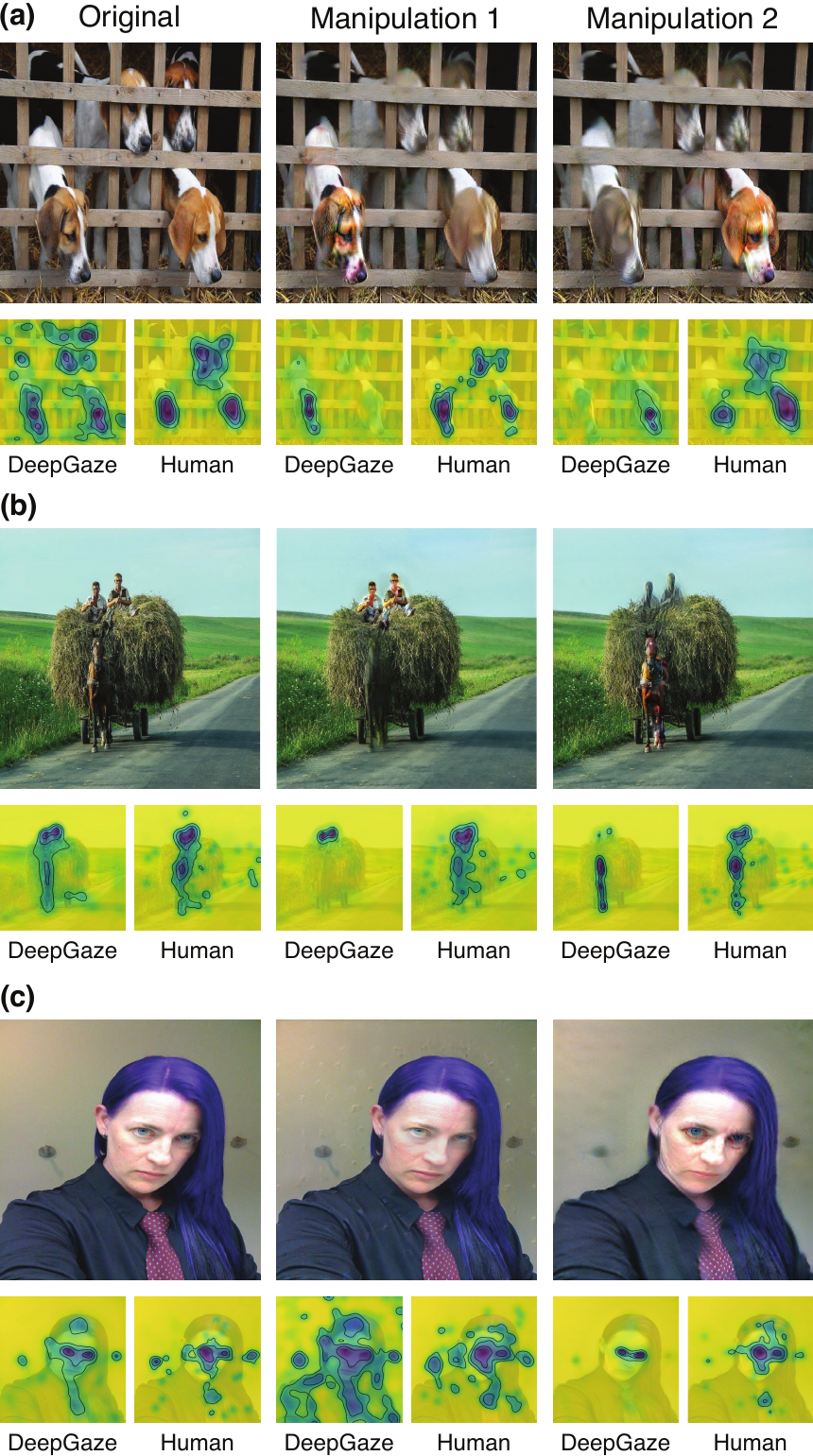}
\caption{ 
Example results of behavioural study.
\textbf{(a)} The image is manipulated to highlight either the dog in the lower right or lower left of the image. 
Fixation probability on the targeted dog increases by 0.17 (77\%) for the dog in the left and 0.17 (70\%) for the dog in the right. 
Deep Gaze predicts stronger increases of 0.64 (285\%) and 0.54 (274\%) respectively.
\textbf{(b)} The image is manipulated to highlight either two humans on the carriage or the horse pulling it. 
Fixation probability increases on average by 0.02 (13\%) for the humans and 0.15 (66\%) for the horse. 
Deep Gaze predicts stronger increases of 0.15 (105\%) and 0.56 (262\%) respectively.
\textbf{(c)} The image is manipulated to in-/decrease the entropy of the fixation density. The empirical entropy in-/decreases by 0.76/0.18 bits. DeepGaze predicts a stronger in-/decrease of 1.08/1.04 bits.
Images best viewed with zoom.
}
\label{fig:BeRes}
\end{figure}

We picked 24 images from the COCO validation set that were not used during training.
For each image, we created two modified versions leading to a total of 72 images.
We designed the modifications to generate opposite changes in human behaviour compared to the original image.
For 21 images, we aimed to in-/decrease the saliency of different objects in the two versions using a mixture of the local shifting and global scaling of the saliency map (e.g. Fig. \ref{fig:BeRes}(a),(b)).
For three images, we purely scaled the saliency map to in-/decrease its entropy (e.g. Fig. \ref{fig:BeRes}(c)).
Since our image transformation is fast (117 ms on a GTX 1080 GPU), we can manipulate images with respect to the parameters $k_{sh}$ and $k_{sc}$ in an online, interactive fashion.

When applying both local shifting and global scaling, we first transformed the image by the network trained on local shifting and afterwards transformed the output of that manipulation by the network trained on global scaling of the saliency map.
In difference to the training, we did not blur the local shift mask when computing the target saliency as this slightly improved the perceptual quality of the manipulations.
Images of all stimuli, the target saliency map to generate them, their saliency map predicted by DeepGaze and their saliency map measured from human behaviour can be found in the Supplement.

DeepGaze is trained on images that are down-sampled by a factor 2 compared to the size of the images on which the fixation data was collected \cite{kummerer_deep_2015}.
Therefore we needed to up-sample the generated images by the same factor before collecting human fixations.
We used a state-of-the art network for superresolution \cite{sajjadi_enhancenet:_2016} to up-sample the generated and original images of size $512\times512$ to size $1024\times1024$ as this gave better results than bicubic up-sampling.

\subsection{Experimental setup}
We measured fixation responses of 23 subjects to our 72 stimuli in a free-viewing task.
Experimental details can be found in the Appendix.

To obtain an empirical fixation density from the raw fixation data, we fit a kernel density estimate together with a uniform component and a center-bias.
We fit this estimate separately for every image and cross-validate over subjects.

To compute DeepGaze's prediction for each stimulus, we use the center-bias computed from the empirical densities from all other stimuli.
\subsection{Results}
Each stimulus was generated to produce a specific behavioural effect. 
For 21 source images, we generated two manipulated versions whose desired effect was to in-/decrease the probability of looking at certain objects (e.g. Fig. \ref{fig:BeRes}(a)).
After obtaining empirical fixation densities from measuring people's fixation patterns, we use the COCO segmentation masks to compute the probability of people looking at the objects in the image: $p_{obj} = \sum_{x,y}M(x,y)p(x,y)$.
In case of manipulations that aimed to increase the probability of looking at certain objects, we measured an average increase in the probability of looking at objects targeted by the manipulation of 0.09.
Relative to the probability of looking at the corresponding object in the original image this result constitutes an increase of 43\%.
For manipulations that aimed to decrease the probability of looking at certain objects, we measured an average decrease in the probability of looking at objects targeted by the manipulation of 0.04 (22\%).

For three source images, we generated two manipulated versions that aimed to change the entropy of the fixation density (Fig. \ref{fig:BeRes}(c)).
We compute the entropy of the empirical fixation densities as $H = -\sum_{x,y} p(x,y) \log(p(x,y))$.
The fixation densities of images manipulated to in-/decrease the entropy of their saliency map showed an average in-/decrease in entropy by 0.19/0.48 bits (1/3\%).
Although the effect size for this manipulation seems small, it can can considerably change the perception of an image
(Fig. \ref{fig:BeRes}(c)).

We can also compare the empirical fixation densities measured from human behaviour with the prediction produced by our saliency model (Fig. \ref{fig:BeRes}).
We find that DeepGaze usually predicts much stronger effects for the manipulated images than we find from human behaviour.
DeepGaze predicts an in-/decrease in fixation probability for manipulated objects of 0.43/0.15 (142/74\%).
For the entropy manipulation, DeepGaze predicts an average in-/decrease by 0.89/1.32 bits (5/8\%).
The discrepancy between model prediction and human behaviour shows that our image transformation still has an adversarial component to it:
It is more effective in manipulating model perception compared to human perception.

\begin{figure}
\includegraphics[width=1\linewidth]{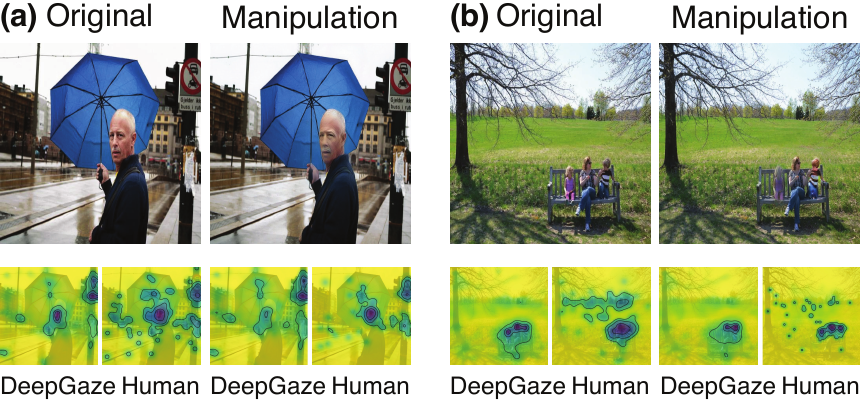}
\caption{ 
Failure examples.
\textbf{(a)} The probability to fixate on the man increases by 0.07 (19\%) although DeepGaze predicts a decrease by 0.32 (44\%).
\textbf{(b)} The probability to fixate on the girl increases by 0.07 (150\%) although DeepGaze predicts a decrease by 0.1 (82\%).
Images best viewed with zoom.
}
\label{fig:FailEx}
\end{figure}

Inspecting the results, we find that the adversarial component is most apparent when human fixations are driven by high-level semantic information.
For example, humans continue to look at a face in the middle of the image, even though the image is manipulated such that DeepGaze assigns considerably less saliency to the face (Fig. \ref{fig:FailEx}(a)).

\section{Discussion}
The ability of machines to predict perceptual properties of images has greatly improved with the rise of CNNs.
Still, it is an open question to what extent these models can inform image manipulation and synthesis with respect to perceptual properties.
Prediction models discard image information that is not informative for their task leading to two major problems when using them for image generation: Ensuring that the output is sufficiently natural and the existence of adversarial examples.

In this work we tackle these problems in the specific case of saliency manipulation.
We use the best model for fixation prediction to learn an image-to-image mapping that can manipulate images to change human fixation patterns in a controlled fashion. 
Nevertheless, we also find some limitations of our saliency manipulation model.
Most prominently, the learned image transformation has difficulties to decrease the saliency of semantically important objects such as human faces.
This can probably only be achieved with severe changes to the semantic content of the image (e.g. deleting a face in some way), which our model has not managed to learn.
It is a compelling question for future work how to improve the design of the transformation network to enable such strong semantic image manipulations.
In summary, we believe this work sets the stage for an exciting new path to edit images and contributes towards enabling the use of powerful prediction models for image manipulation.

\section{Appendix}
\subsection{Training details}
\subsubsection{Transformer network}
We set the weights of the loss function to $\lambda_{s}=\num{1e0}, \lambda_{f}=\num{1e-2}, \lambda_{t}=\num{2e-2}, \lambda_{a}=\num{1e-1}$ after a preliminary exploration phase on smaller datasets.
We trained separate network instance for each type of saliency manipulation, the local shifting and the global scaling.
Every network was trained for 150k iterations with batch size 4 and learning rate $\num{1e-3}$ using the Adam optimiser \cite{kingma_adam:_2014}.
As a comparison to our FGTransform, we also trained instances of the `Resnet-9' and `U-Net' architectures from \cite{johnson_perceptual_2016, isola_image--image_2016} with two slight modifications: We replaced BatchNorm by InstanceNorm and we initialised them as an auto-encoder by first training them to reconstruct the input image, which slightly accelerated training in the beginning.
We also experimented with  the `context-aggregation network' (CAN32) from \cite{chen_fast_2017} but did not get promising results in our task.

\subsubsection{Adversarial loss}
The discriminator in the adversarial loss is a five-layer CNN with 64, 128, 256, 512 and 1 channels and LeakyReLU non-linearity.
In the last layer, each unit has receptive field size of 70px and is trained to discriminate between transformed and real image patches.
In difference to \cite{isola_image--image_2016} we used InstanceNorm instead of BatchNorm layers in the discriminator.
To implement the adversarial loss, we slightly modified the code\footnote{github.com/junyanz/pytorch-CycleGAN-and-pix2pix} from \cite{isola_image--image_2016, zhu_unpaired_2017}.

\subsubsection{Extended results}
\begin{figure}
\includegraphics[width=1\linewidth]{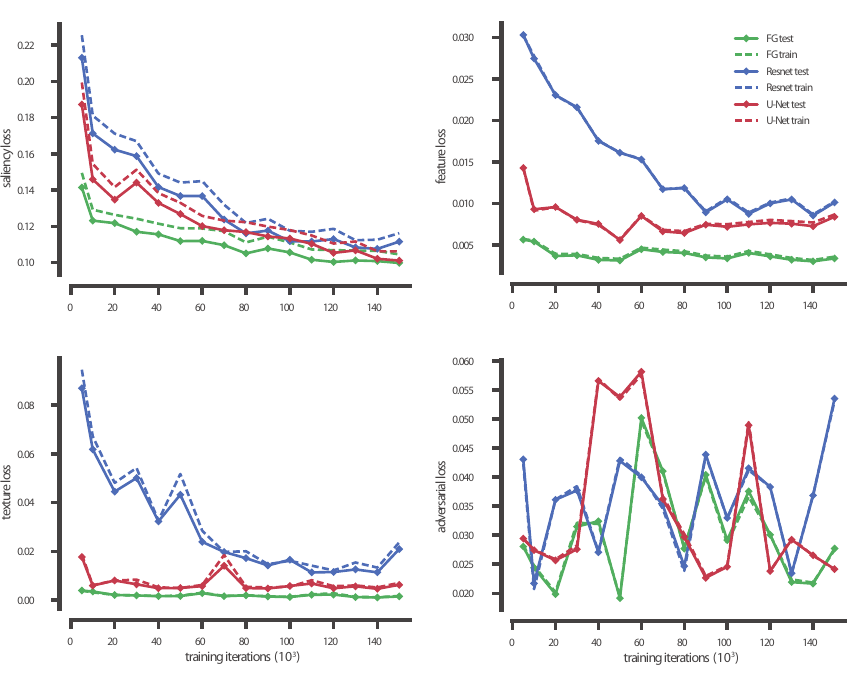}
\caption{
Quantitative evaluation of network trained with target saliency maps generated by global scaling of the original saliency map.}
\label{fig:SuppTrain}
\end{figure}

In Fig. \ref{fig:SuppTrain} we show the training and test loss for the networks trained with target saliency maps from global scaling of the original saliency maps.
The Ablation and Model Comparison results for all stimuli used in the behavioural experiment can be found in the \textbf{`ModelComparison'} and \textbf{`AblationStudies'} folders in the Supplement at: \\\emph{bethgelab.org/media/uploads/gazeguide/Supplement.zip}.

\subsection{Behavioural study}
23 participants were recruited from an internal mailing list. 
Each participant saw three blocks of 24 stimuli; each block contained one version of the 24 source images (one block with the original images and two blocks with manipulated images).
The order of the blocks was counterbalanced over participants (latin square design) such that an approximately equal number of participants saw each condition in their first block.
On each trial, participants were presented with a fixation target in the centre of the screen.
After fixating, the images were displayed for 3 seconds and could be freely explored by the participants.
The inter-trial interval was 2 seconds, in which a blank grey screen was presented.

Stimuli were displayed on a VIEWPixx 3D LCD (VPIXX Technologies Inc., Saint-Bruno-de-Montarville, Canada; spatial resolution 1920 × 1080 pixels, temporal resolution 120 Hz). 
Participants viewed the display from 60 cm in a darkened chamber. 
At this distance the images subtended approximately 25 degrees of angle at the retina.
Stimuli were presented using the Psychtoolbox Library \cite{brainard_psychophysics_1997, pelli_videotoolbox_1997, kleiner_whats_2007} version 3.0.12 under MATLAB (The Mathworks Inc., Natick MA, USA; R2015b).
Participants' gaze position was recorded monocularly (left eye) at 500 Hz with an Eyelink 1000 (SR Research, Ontario, Canada) video-based eyetracker in combination with the Eyelink toolbox \cite{cornelissen_eyelink_2002} for MATLAB.
Gaze traces were classified into fixations using the default settings of the SR Research processing software.

There are two potential problems in the data.
First, every participant saw each image 3 times (once per block). 
Thus, memory effects potentially changed behaviour.
We controlled for this by counterbalancing the order of the blocks over participants to minimise the influence of memory / familiarity on the average densities over conditions.
We additionally analysed the data of only the first block for each participant and found a similar effect in human behaviour as for the full data (average in-/decrease to fixate the targeted objects: 0.08/0.03 (55/9\%)).

\begin{figure}
\includegraphics[width=1\linewidth]{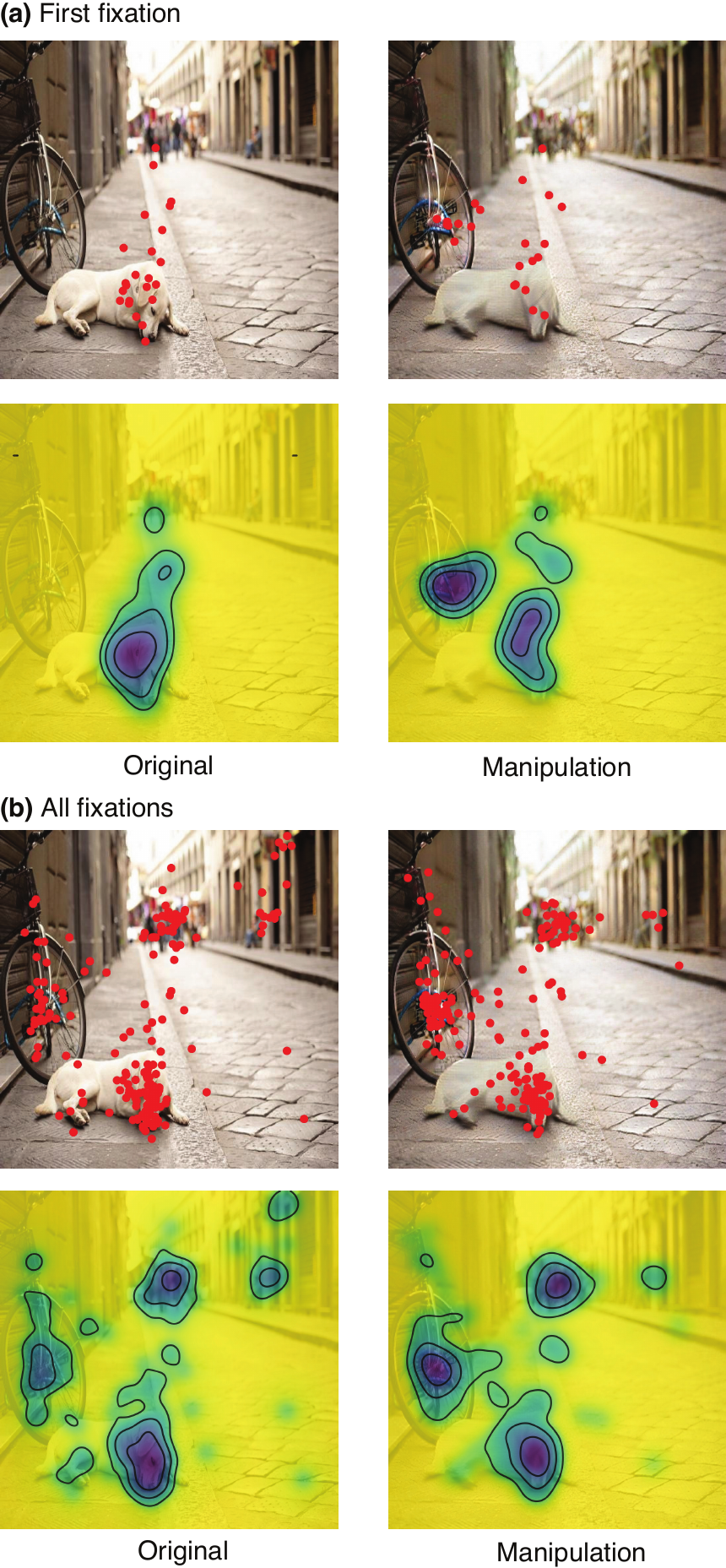}
\caption{
\textbf{(a)} Raw fixation data and estimated empirical density for only the first fixation for all subjects. There are no fixations on the bicycle on the left for the original but several for the manipulated image making the relative increase in fixation probability extreme (0.31/111648\%).
\textbf{(b)} Raw fixation data and estimated empirical density for all fixations from all subjects.
Fixations are more distributed and the relative increase of fixation probability on the bicycle is therefore less extreme (0.12/70\%).
}
\label{fig:SuppFF}
\end{figure}

Second, although all images except for one (the snowboarder) were unseen by all participants before the experiment, some participants were familiar with the general research question.
Thus, fixations could be influenced by conscious behaviour.
To check if this could potentially invalidate our results we also analysed only the first fixations from all participants, under the assumption that the first fixation is difficult to control voluntarily. 
We found that for this subset of the data the effect size rather increased (average in-/decrease to fixate the targeted objects: 0.15/0.10 (3230/35\%)).
The reason for the very large average relative increase is that for some manipulated images, a significant amount of first fixations is guided to target objects that received no first fixation in the original image (e.g. see Fig. \ref{fig:SuppFF}, fixations on the bicycle). 
This generates a huge relative increase in fixation probability for these images, which dominate the average value.
The median relative in-/decrease in fixation probability for the first fixation only was 57/48\%.

All data from the experiment is contained in the folder \textbf{`BehaviouralExperiment'} in the Supplement at: \\\emph{bethgelab.org/media/uploads/gazeguide/Supplement.zip}:
\begin{itemize}
\item \textbf{`BehaviouralExperiment/Stimuli'} contains the images shown in the experiment 
\item \textbf{`BehaviouralExperiment/TargetSaliencyMaps'} contains the target saliency maps used to produce the stimuli. Note that the target saliency maps can look artificial, since we did not use blurring of the object masks as we did during training. Training directly with non-blurred object masks did not improve the results.
\item \textbf{`BehaviouralExperiment/DeepGazePrediction'} contains the saliency prediction by DeepGaze in response to the stimuli images. The difference between the target saliency and the saliency prediction is quantified by the test loss of the model (although the loss was again evaluated with blurred object masks).
\item \textbf{`BehaviouralExperiment/HumanFixations'} contains the human fixation data in response to the stimuli in the folder.
\item \textbf{`BehaviouralExperiment/DataOnlyFirstBlock'} contains the human fixations and DeepGaze predictions for only the data in the first block for each subject. Note that DeepGaze predictions are slightly different, because the center-bias is also adapted to contain only the data from the first block.
\item \textbf{`BehaviouralExperiment/DataOnlyFirstFixation'} contains the human fixations and DeepGaze predictions for only first fixation of all subjects and images. Note that DeepGaze predictions are slightly different, because the center-bias is also adapted to contain only the data from the first fixation.
\end{itemize}

{\small
\bibliographystyle{ieee}
\bibliography{SaliencyManipulation}
}

\end{document}